# LISBET: a machine learning model for the automatic segmentation of social behavior motifs

Giuseppe Chindemi*, Benoit Girard* and Camilla Bellone
Department of Basic Neurosciences, University of Geneva, Switzerland

*equal contribution

Corresponding authors: camilla.bellone@unige.ch; giuseppe.chindemi@unige.ch

## Abstract

Social behavior is crucial for survival in many animal species, and a heavily investigated research subject. Current analysis methods generally rely on measuring animal interaction time or annotating predefined behaviors. However, these approaches are time consuming, human biased, and can fail to capture subtle behaviors.
Here we introduce LISBET (LISBET Is a Social BEhavior Transformer), a machine learning model for detecting and segmenting social interactions. Using self-supervised learning on body tracking data, our model eliminates the need for extensive human annotation.
We tested LISBET in three scenarios across multiple datasets in mice: supervised behavior classification, unsupervised motifs segmentation, and unsupervised animal phenotyping. Additionally, in vivo electrophysiology revealed distinct neural signatures in the Ventral Tegmental Area corresponding to motifs identified by our model. In summary, LISBET automates data annotation and reduces human bias in social behavior research, offering a promising approach to enhance our understanding of behavior and its neural correlates.

## Introduction

Animal behavior research has traditionally relied on human observers to categorize and label stereotypical movements or actions. While this approach has yielded valuable insights, it presents significant challenges that have long constrained the field. In particular, manual annotation of behaviors for quantitative research is often tedious, subjective, and prone to errors, potentially compromising the reliability and reproducibility of studies. Furthermore, the specific set of behaviors chosen for analysis in a study inherently introduces researcher bias, potentially limiting the scope of discoveries.

Recent advancements in machine learning methods within biological disciplines are addressing these challenges and revolutionizing behavioral research. These algorithmic approaches are enhancing our understanding of animal behavior in two important ways. On one hand, machine learning can replicate and automate human annotations through a hypothesis-driven approach, mitigating the issues of reproducibility and human error (Bohnslav *et al.*, 2021; Marks *et al.*, 2022). This application not only increases the efficiency of data analysis but also improves the consistency and reliability of behavioral classifications. On the other hand, these computational methods enable the discovery of new behavioral motifs through a data-driven approach, potentially revealing patterns that human observers might overlook (Wiltschko *et al.*, 2020; Dunn *et al.*, 2021; Hsu and Yttri, 2021; Luxem *et al.*, 2022). This capability expands the breadth of behavioral investigations beyond preconceived notions, offering opportunities to uncover previously unrecognized aspects of animal behavior. By leveraging these machine learning techniques, researchers can simultaneously enhance the efficiency and objectivity of behavioral analysis while also expanding the scope of their investigations.

However, despite the promising results obtained by these new tools in behavioral research, most progress in the field has been made in single-animal settings. Social interactions, characterized by the interplay of multiple individuals, introduce greater challenges that are not easily addressed by simply extending single-animal methods. While single-animal actions can be deconstructed into simple features like velocity (i.e., locomotion) or body part positions (i.e., rearing), social interactions span multiple timescales and can be of extreme complexity (e.g., group hunting). Classification approaches based on heuristics or supervised learning methods have shown promising results in specific social scenarios, such as physical contact or chasing (de Chaumont *et al.*, 2019; Segalin *et al.*, 2021; Marks *et al.*, 2022; Ye *et al.*, 2023; Goodwin *et al.*, 2024). However, these methods still offer limited generalization to novel conditions and introduce anthropomorphic bias that may fail to capture the extensive spectrum of social behaviors exhibited by animals.

The success of unsupervised methods in single-animal studies has naturally led to considerations of their potential applicability in social behavior analysis. One example of this approach is Keypoint MoSeq (Weinreb *et al.*, 2024), which excels in identifying behavioral motifs in solitary animals and has been applied to social behavior settings in dyadic interactions. This method operates on the premise that observing one animal can often provide sufficient information to infer the nature of the social interaction between the two animals. However, this approach is inherently limited to a subset of behaviors that exhibit this reciprocal property and does not readily extend to social contexts involving larger numbers of individuals. Moreover, the extension of single-animal methods to multi-animal scenarios presents fundamental challenges. Social behaviors typically span multiple timescales, including very long ones, making it impractical to train models on predicting future positions of animals, an approach often used in unsupervised single-animal tools.

Unsupervised methods specifically designed for social behavior research are in their infancy. Recent approaches such as SBEA (Han *et al.*, 2024) and DeepOF (Miranda *et al.*, 2023) offer promising solutions to unsupervised behavior analysis for multiple animals, but may struggle with capturing the nuanced and dynamic nature of social interactions across multiple timescales. Furthermore, these tools have not yet been benchmarked against publicly available datasets or demonstrated their neurobiological applicability.

To address these challenges, we have developed LISBET (LISBET Is a Social BEhavior Transformer), a machine learning model designed for the automated discovery and classification of social behavior motifs. LISBET analyzes body point coordinates obtained from video recordings of mice pairs, leveraging computational methods to uncover patterns in social interactions. LISBET is built upon the ViT/ViViT architectures (Dosovitskiy *et al.*, 2020; Arnab *et al.*, 2021). The model is trained to generate a high-dimensional embedding of the observed scene by simultaneously solving four self-supervised learning tasks, drawing inspiration from the literature on large language models (Devlin *et al.*, 2019). This approach allows LISBET to capture the nuanced complexities of social behaviors without relying on pre-defined categories or extensive labeled datasets. LISBET processes body point coordinates of mice, which can be obtained from various pose estimation tracking software such as DeepLabCut (Mathis *et al.*, 2018), MARS (Segalin *et al.*, 2021) or SLEAP (Pereira *et al.*, 2020), facilitating its integration into existing research workflows across different experimental setups.

LISBET is designed to operate in two complementary modes for social behavior analysis. In the hypothesis-driven mode, LISBET aims to extract key features of social behaviors and classify them in a manner that mirrors human annotations. In the discovery-driven mode, LISBET detects and segments social behavior motifs without prior examples, allowing for the identification of potentially novel or overlooked behavioral patterns. Our findings demonstrate the effectiveness of both operational modes. In the hypothesis-driven approach, LISBET successfully classifies behaviors in close alignment with human annotations, showing its potential to automate and enhance traditional behavioral scoring methods. In the discovery-driven approach, LISBET identifies meaningful behavioral motifs without prior examples. Importantly, the motifs identified through LISBET's discovery-driven approach not only align closely with human annotations but also correlate with the electrophysiological activity of Ventral Tegmental Area (VTA) dopaminergic neurons recorded in vivo in freely moving animals. This correlation between computationally derived behavioral motifs and neurophysiological data provides crucial validation of LISBET's biological relevance, bridging the gap between machine learning-based behavior analysis and neurobiological understanding.

LISBET is provided to the community as a Python tool to automate social behavior annotation, subjects stratification, and to study novel experimental conditions hypothesis-free. Code and weights of the best-performing models are available in closed beta testing upon request to the authors, and will be made publicly available in the upcoming weeks at https://github.com/BelloneLab/lisbet.

# Results

LISBET is a model designed to analyze video recordings of social behavior in animal experiments (Fig. 1). The model operates on coordinates of animal body parts (Mathis *et al.*, 2018; Pereira *et al.*, 2020; Pereira, Shaevitz and Murthy, 2020; Dunn *et al.*, 2021; Segalin *et al.*, 2021) using a sliding window over a video. Each window is processed by a transformer encoder, based on the ViT/ViViT transformer architectures (Dosovitskiy *et al.*, 2020; Arnab *et al.*, 2021), to generate an embedding (i.e., a feature vector that encodes a compressed representation of the window). The embedding can be used to train a model on a small subset of labeled data to automate the annotation of a large set of recordings, as traditionally done in supervised learning (Fig. 1 "hypothesis driven mode"). Alternatively, a clustering algorithm can be used to segment embeddings into motifs, akin to human-defined behaviors, to study social interactions without a human-defined set of labels (Fig. 1, "discovery driven mode"). In the following sections, we described these two modes and provide concrete examples of their applications.

## LISBET automates the annotation of human-defined behaviors

Transformer models such as our encoder (Fig. 1) are usually pre-trained on large unlabeled datasets, before being fine-tuned for a specific application using a smaller labeled dataset (Devlin *et al.*, 2019). This approach is particularly well suited for our purposes, as it allows us to expose the model to a large body of tracking data without biasing it with human annotations until requested. That is, the pre-trained encoder can be used as is to segment tracking data into behavioral motifs (discovery-driven mode), or fine-tuned using human labels to identify a desired set of behaviors (hypothesis-driven mode).

To pre-train the LISBET encoder we used the unlabeled tracking dataset provided in CalMS21 (Sun *et al.*, 2021). In brief, this dataset contains the coordinates of seven body parts of two mice during a resident-intruder paradigm. To guide the encoder to produce useful embeddings for social behavior analysis, we introduced a self-supervised approach emphasizing the most salient aspects of social interactions. This method requires the model to distinguish between genuine and manipulated data of social interactions in artificially generated scenarios (i.e., tasks, Fig. 2a). The first task, "Swap Mouse Prediction (SMP)", requires the model to predict whether the given input window is an authentic video segment or an artifact generated by selecting the body part coordinates of two mice from two different videos. The second task, "Next Window Prediction (NWP)", requires the model to distinguish whether two successive input windows are extracted from the same video or whether the second one is randomly sampled from the dataset. The third task, "Video Speed Prediction (VSP)", requires the model to predict whether the sampling rate of an input window corresponds to the original one. Finally, the fourth task, "Delay Mouse Prediction (DMP)", requires the model to determine whether the mice presented in an input window are aligned in time or artificially delayed one another.

After pre-training on the four self-supervised tasks (Extended Data Fig. 1), we visually inspected the embeddings produced by the encoder using the test set of the CalMS21 Task 1 dataset (Sun *et al.*, 2021). This dataset contains annotations for three common social behaviors (i.e., attack, investigation and mount) and was labeled by a single person. Overlaying the human annotations onto the embeddings revealed that distinct behaviors tended to occupy specific regions of the embedding space, despite the fact that neither human labels nor the actual tracking data from this dataset were used during pre-training (Fig. 2b). This result indicates that the embeddings might be used as is to segment motifs matching human-annotated behaviors (see Section "Social behavior motifs discovery using LISBET embeddings" below). Furthermore, it suggests that the encoder obtained after pre-training could be a good starting point for fine-tuning a model to reproduce specific human annotations.

To test the latter hypothesis, we trained a classifier to predict the human-annotated behaviors in the CalMS21 Task 1 dataset (Sun *et al.*, 2021) from the encoder's embeddings of the data (Fig. 1 "hypothesis driven mode"). The model obtained by freezing the encoder weights while optimizing the classifier obtained an F1 score of 0.70, confirming the intuition that the pre-trained embeddings are already informative of social behavior (Fig. 3a). Allowing the optimizer to also update the encoder weights raised the test F1 score to 0.79 (Fig. 3b), and removing the constraint of purely causal windows allowed the model to obtain a test F1 score of 0.81 (Fig. 3c). For reference, the best performing model for this dataset reported a test F1 score of 0.86 (Sun *et al.*, 2021), although a fair comparison would require testing the two model in similar conditions (i.e., the best model was pre-trained/trained on data from all available splits/tasks). Finally, fitting the model without pre-training the encoder achieved an F1-score of 0.73, highlighting the importance of self-supervised pre-training to obtain a good classifier. The results of all our tests on the CalMS21 Task 1 dataset (Sun *et al.*, 2021) are summarized in Table 1.

In conclusion, LISBET successfully extracts features of social interaction without the need for human annotations. Furthermore, after fine-tuning, the model can automate human annotation.

## LISBET discovers social behavior motifs without human supervision

To test whether the pre-trained encoder could be used as is to segment social behavior into motifs, without fine-tuning on human annotations, we fitted a Hidden Markov Model (HMM) (Murphy, 2012) on the embeddings obtained from the CalMS21 Task 1 dataset (Fig. 4a). This approach has already been proved successful to segment single animal behavior (Wiltschko *et al.*, 2020). A HMM is a statistical model, where the observations are obtained by sampling from a set of hidden states (i.e., random distributions), whose parameters are estimated during fitting to maximize the probability of observing the data. In this context, the embeddings are used as observations and hidden states correspond to the behaviors inferred from the embeddings. After fitting the HMM, we visually inspected the motifs produced by the model and observed a qualitative agreement between motifs and the human annotations (compare Fig. 4a and Fig. 2b), suggesting that the HMM can identify behaviorally relevant motifs, without human supervision. This agreement extends to the temporal dynamics of the behaviors (Fig. 4b).

Unsupervised learning algorithms often require specifying the number of classes a priori (i.e. the number of motifs), which can lead to over- or under-categorization. To address this issue, we used a hierarchical clustering approach to group motifs from different HMMs, based on their Jaccard distance as a measure of similarity. That is, motifs repeatedly found by different HMMs are grouped together in a macro-motif. The number of macro-motifs is determined by maximizing the average silhouette score of the hierarchical clustering. As the motifs in a macro-motif are by definition similar, rather than analyzing all of them we can identify a prototype motif representative of the group. We considered as the prototype the motif with the highest silhouette score within the macro-motif (Fig. 4c). Using this approach allowed us to specify the maximum number of motifs, which can be easily determined by imposing a limit on the granularity of the HMM segmentation, rather than deciding a priori the exact number of motifs.

We compared the prototype motifs with the human annotation using the F1 score and found that several prototypes overlay with human-identified behaviors (Fig. 4d, NMI = 0.32). To our knowledge, the best previously reported NMI value for this dataset by unsupervised tools was below 0.07 (Weinreb *et al.*, 2024). Closer inspection of the results revealed that the same human-identified behavior can be represented by multiple prototypes. For example, our algorithm produced two motifs, prototype 7 (p7) and prototype 8 (p8), that would be considered as instances of "investigation" by a human (see Supplementary Video 1 for a random sample of p7 events, and Supplementary Video 2 for p8). To investigate why the algorithm divided these two prototypes, we compared the mean bout duration and frequency of these two motifs, but we did not find any significant differences (Fig. 4e,f). Since social behavior is a dynamic process, we hypothesized that the meaning of the two prototypes could be better understood by looking into longer behavioral sequences, composed by multiple interactions. To test that, we built a prototype transition matrix and found that p7 and p8 belong to a distinct loop (p7 to p8 and vice versa), terminated by a transition from p8 to the non social state p2 (Fig. 4g, see Supplementary Video 3 for a random sample of p2 events).

Our results show that LISBET can be used to discover behaviorally relevant motifs, without human supervision. These motifs can be characterized in terms of mean duration, rate and transition probabilities.

## LISBET automates social behavior phenotyping

Another interesting application of our model is behavioral phenotyping (e.g., comparing different mouse lines or experimental groups). To showcase this approach in the discovery-driven mode, we compared the behavior of male mice when exposed to a mouse of the same versus opposite sex. We processed video recordings from the CRIM13 dataset (Burgos-Artizzu *et al.*, 2012), extracted body point coordinates using DeepLabCut (Mathis *et al.*, 2018) and generated social behavior embeddings with LISBET. Analyzing LISBET embeddings using UMAP revealed distinct clusters for male and female intruders (Fig. 5a). Subsequently, we applied the same approach as in Fig. 4 to segment these features into motifs and select their prototypes (Fig. 5b). To confirm the behavioral relevance of the prototypes, as before we compared them with the human annotations available for this dataset (Fig. 5c). Among the 14 prototypes, we found that p2 and p3/p4 overlay with "attack" and "copulation" respectively. These two behaviors are usually found to be more expressed in males and females respectively, and further analysis of the duration and rate of motif bouts showed significant differences between the groups (Fig. 5d,e). This result confirmed that prototypes, discovered automatically by LISBET without any human supervision, are good predictors of sex-specific behavior. Extending the analysis of prototypes transition described in Fig. 4 to characterize the behavioral strategies of two distinct groups, we found that the interactions with a male and a female can also be distinguished by their transition dynamics (Fig. 5f). For example, the male to male interaction is characterized by the p0 to p2 loop (Fig. 5f right, blue arrows), while the male to female interactions are more commonly expressed as the p13 to p3 or p2 to p3 transitions (Fig. 5f right, orange arrows).

These findings demonstrate how LISBET can be used in a discovery-driven mode to characterize social phenotypes without relying on prior assumptions about the nature of the group-specific behaviors. Furthermore, it exemplifies how LISBET can provide new ethological insights by looking at the differences in transition dynamics.

## LISBET discovers neuronal correlates of social behavior

As LISBET motifs do not depend on human interpretation, we hypothesized that they could reveal the neural correlates of social interactions beyond the limited set of stereotypical behaviors commonly used in literature.

To test this hypothesis, we tracked freely moving mice in dyadic interaction using DeepLabCut (Mathis *et al.*, 2018), while concurrently recording spike unit activities of putative dopaminergic neurons in the Ventral Tegmental Area (VTA-pDA neurons) (Fig. 6a, left). Then, using LISBET, we identified the motif prototypes as in Fig. 4 (Extended Data Fig. 2). We found that several prototypes were highly correlated with neuronal activity of the VTA-pDA, with diverging patterns (Fig. 6a, right). In particular, p19 corresponded to an increase in activity, while p0 corresponded to a decrease. For reference, some prototypes as p10 were not correlated with neuronal activity. Interestingly, looking at the corresponding embeddings for these prototypes using UMAP showed that p0 and p19 are segregated one from another, while p10 was scattered in the embedding space (Fig. 6b).

Video analysis of these motifs revealed subtle differences, not easily associated with a canonical behavior (see Supplementary Video 4, 5, and 6 for a random sample of p0, p10, and p19 events respectively). This result is largely expected, as detecting subtle and unexpected behaviors is the main objective of LISBET's discovery-driven mode. To quantify these different prototypes, and help our intuition on the nature of the behaviors, we measured three key metrics from the corresponding video segments (Fig. 6 c-e). We found that p19 is characterized by a quick approach of the experimental mouse, often followed by an escape behavior of the stimulus mouse, consistent with a behavior we could describe as "unilateral forced investigation toward the stimuli". On the other hand, p0 is characterized by a hesitant behavior, also not reciprocated by the experimental mouse, consistent with a behavior we could describe as "hesitant investigation".

Taken together, our results show that LISBET-derived prototypes not only match human annotations, but also allow the identification of rare and subtle behaviors aligning with specific neuronal activities.

# Discussion

In this work we introduced LISBET, a machine learning pipeline for social behavior segmentation and classification. We have shown that training the model using four self-supervised tasks produces generalizable embeddings suitable for behavior classification and phenotyping with no fine-tuning required. We also described our approach to dynamically classify and segment data at various granularity levels, selecting optimal macro-categories of Motifs and identifying a prototype for each. Finally, we used LISBET to correlate the neural activity of VTA-pDA neurons with the predicted behavioral motifs. Interestingly, we found that a few distinct prototypes presented unique neuronal signatures despite being virtually indistinguishable using traditional hypothesis-driven methods. Taken together, these findings suggest that LISBET can be used to expand the analysis toolkit of behavioral neuroscientists, minimizing the need for human-labeled data and drastically reducing the impact of the corresponding biases on the results.

A key strength of LISBET is its use of transformer architecture (Vaswani *et al.*, 2017; Dosovitskiy *et al.*, 2020; Arnab *et al.*, 2021), which allows it to ingest and learn from large quantities of data (Kaplan *et al.*, 2020). For this reason, LISBET's performance is expected to continually improve as more behavioral data becomes available. Furthermore, the convergence of representations in machine learning suggests that large foundational models, including those trained on natural language processing tasks, could be adapted for behavioral analysis (Huh *et al.*, 2024). This opens up exciting possibilities for leveraging pre-existing models and knowledge to enhance LISBET's capabilities.

The self-supervised tasks used to train LISBET were inspired by literature on natural language processing (Devlin *et al.*, 2019), and carefully designed to capture key aspects of social behavior, such as spatial relationships, temporal patterns, speed, and synchronicity. These tasks enable the model to learn meaningful representations of social interactions without relying on human labels, making it a powerful tool for discovery-driven research. This approach also allows the model to learn from large amounts of unlabeled data and reduces the need for extensive manual annotation, a significant bottleneck in traditional behavioral analysis pipelines.

While LISBET can effectively be used to automate data annotation and obtained comparable, although slightly lower, results than the state of the art methods on the CalMS21 dataset (Sun *et al.*, 2021), it is important to note that our primary goal was not to optimize LISBET for the CalMS21 benchmark specifically. Rather, we wanted to develop a versatile and generalizable model for social behavior analysis. For example, we only use past frames for predictions to make the model suitable for neuronal activity analysis and close-loop circuit investigation. The choices made in designing LISBET, such as the window size of 200 frames, were based on intuition and iterative testing, but these parameters can be easily adjusted to suit specific experimental needs.

One of the most compelling aspects of LISBET is its ability to discover motifs without human supervision using HMMs (Murphy, 2012). While these motifs may not perfectly reproduce human annotations, they have the potential to identify behaviors more objectively, as they are not influenced by individual human biases. HMM-based segmentation allows LISBET to uncover subtle or previously unrecognized behaviors that may be overlooked by human annotation. In fact, it has the potential to surpass human performance in identifying behaviors, as it doesn't capture the biases or preferences of a single human annotator. To test this hypothesis, we could compare the annotations produced by LISBET with the ensemble-human-annotations, obtained by averaging the choices of multiple expert human annotators on the same dataset.

We demonstrated LISBET's suitability for behavioral phenotyping by comparing male-male and male-female interactions, but the model's applications extend far beyond this example. LISBET could be used to study a wide range of conditions, including neuropsychiatric disorders, by identifying unique behavioral patterns associated with these conditions. Moreover, while our current study focused on dyadic interactions, LISBET could be extended to analyze group behaviors by composing dyadic interactions or by developing models that natively support the analysis of complex group dynamics, such as cooperation or social hierarchy formation. This versatility makes LISBET a valuable tool for researchers across multiple domains of neuroscience and behavioral biology.

Our finding that LISBET motifs correlate with VTA-pDA neuronal activity highlights the model's potential for uncovering the neural underpinnings of social behavior. By objectively segmenting behavior and aligning it with neural recordings, LISBET can help researchers identify the specific neural circuits and mechanisms that drive different aspects of social interaction. Our discovery of distinct motifs in VTA activity during social interactions highlights the power of LISBET in uncovering neural correlates of behavior. This finding not only validates the biological relevance of our computationally derived behavioral segments but also opens new avenues for investigating the neural basis of social behavior. Future studies could leverage this approach to explore how different brain regions encode various aspects of social interactions, potentially leading to a more comprehensive understanding of the neural circuits underlying social behavior. Looking forward, the integration of LISBET with other cutting-edge neuroscience techniques, such as optogenetics or large-scale neuronal recording like neuropixel or calcium imaging, could provide unprecedented insights into the causal relationships between neural activity and social behavior. Additionally, applying this model to different species or in various environmental contexts could help elucidate the evolutionary and ecological aspects of social behavior.

To contextualize LISBET's contributions, it is important to compare its capabilities with existing tools for rodent behavior analysis. In the domain of supervised behavior analysis, tools such as DeepEthogram (Bohnslav *et al.*, 2021), Live Mouse Tracker (de Chaumont *et al.*, 2019), SiMBA (Goodwin *et al.*, 2024), and MARS (Segalin *et al.*, 2021) have demonstrated high accuracy in replicating human annotations for social behavior analysis in rodents. These approaches, while effective, are constrained by the availability and quality of labeled data and may inadvertently perpetuate biases present in human annotations. LISBET addresses these challenges through its self-supervised learning approach, which reduces dependence on extensive human-annotated data and potentially mitigates inherent biases. While LISBET performs competently with labeled data, it particularly excels in scenarios with limited labeled data but abundant unlabeled tracking data.

Several unsupervised tools designed for single-animal behavior analysis, such as B-SOiD (Hsu and Yttri, 2021), VAME (Luxem *et al.*, 2022), and Keypoint MoSeq (Weinreb *et al.*, 2024), have shown efficacy in their intended contexts and some have been applied in multi-animal settings. However, the extension of these tools to social behavior analysis presents significant challenges, as they may not fully capture the complexity of multi-animal interactions. LISBET, specifically designed for social behavior analysis, addresses these limitations through its architecture and training approach tailored to multi-animal scenarios.

In the field of unsupervised social behavior analysis, recent tools such as SBEA (Han *et al.*, 2024) and DeepOF (Miranda *et al.*, 2023) have shown promising results. While these tools share a similar scope with LISBET, they are built around fundamentally different philosophies. LISBET's distinctive approach is centered on its self-supervised learning procedure, which sets it apart from other tools in this domain. This training method allows LISBET to extract meaningful features from unlabeled data, potentially capturing a broader range of patterns tailored toward social behaviors. Furthermore, LISBET offers multi-granularity behavior detection through hierarchical clustering, enabling the exploration of behaviors at various levels of detail. Notably, LISBET has demonstrated validated alignment with neural activity data, providing a crucial link between computational behavior analysis and neurophysiology, and has been tested against publicly available datasets such as CalMS21 (Sun *et al.*, 2021) and CRIM13 (Burgos-Artizzu *et al.*, 2012). These features collectively enhance researchers' ability to explore social behaviors comprehensively and to establish connections between behavioral patterns and neural dynamics.

While LISBET offers these advanced capabilities, it is important to acknowledge that the choice of tool should be guided by the specific requirements of each research project. For instance, some approaches may provide superior usability in certain contexts, particularly those that integrate animal tracking and behavior analysis. Nonetheless, LISBET's strengths in handling complex social interactions, its dual-mode operation, and its demonstrated biological relevance make it a valuable addition to the toolkit for researchers studying social behavior in rodents. It is particularly suited for scenarios involving complex, multi-animal interactions or when exploring novel behavioral patterns.

However, while our findings are promising, this study has several limitations. First, we only considered pair interactions in our study. Our model is not bound to a given number of animals, but extending it to include more than two animals is beyond the scope of this work. Furthermore, we chose the set of self-supervised training tasks based on intuition and successive iterations. While we were careful not to use the CalMS21 - Task 1 data for self-supervised training, the generalization power to this dataset was implicitly used as the success measure of the training procedure. For this reason, the self-supervised training tasks should not be considered as the absolute best choice for any possible mouse behavior but as an educated guess of what to look for in a video to recognize at least the most commonly investigated mouse behaviors. That is, other behaviors might have been better predicted using other self-supervised training tasks, emphasizing different aspects of mouse behavior. However, our results show that the model can generalize to other datasets and behaviors (i.e. CRIM13, and our in-house dataset), supporting the choice of these tasks. Second, in this study, we did not consider complex social scenarios or species other than mice. We are currently investigating extensions of LISBET applied to human behavior, but research is still at an early stage. Third, while we found initial evidence of behavioral motifs correlation with neuronal activity, more research will be required to delve deeper into the precise neurological pathways and mechanisms that dictate these behaviors. Last, our data pipeline is composed of three sequential steps, namely body-pose estimation, LISBET embedding, and HMM segmentation. Ideally, these steps could be concatenated to produce a more convenient end-to-end solution, lowering the burden on the final user. Furthermore, LISBET is a young project and educational resources on how to use it are still under active development. These and other usability improvements are being deployed over time and we are committed to supporting LISBET for the foreseeable future.

In conclusion, LISBET represents a significant advance in the field of social behavior analysis, offering a data-driven, unbiased, and flexible approach to behavioral segmentation and classification. As the model continues to evolve and be applied to diverse datasets, we anticipate that it will yield novel insights into the complex nature of social behavior and its neural substrates, potentially informing new approaches to studying and treating neuropsychiatric disorders characterized by social deficits.

# Methods

## Software and Tools

LISBET was developed in Python (Van Rossum and Drake Jr, 1995), using TensorFlow (Martín Abadi *et al.*, 2015) and Keras (Chollet and others, 2015). The Hidden Markov Models (HMMs) were fitted using hmmlearn (hmmlearn, 2014). Dimensionality reduction for the visualization of the LISBET embeddings was performed using UMAP via umap-learn (McInnes, Healy and Melville, 2020). Standard scientific Python libraries were employed for data analysis, processing and visualization: numpy (Harris *et al.*, 2020), scipy (Virtanen *et al.*, 2020), pandas (McKinney, 2010), matplotlib (Hunter, 2007) and scikit-learn (Pedregosa *et al.*, 2011). Software development and data analysis were performed in the jupyterlab environment (Kluyver *et al.*, 2016).

## Datasets

The CalMS21 dataset (Sun *et al.*, 2021) contains over one million frames from tracked videos, divided into three different annotation groups for the tasks of the MaBE 2021 competition: classic frame classification (Task 1), annotation style transfer (Task 2), and few-shot learning (Task 3). Furthermore, the dataset contains a set of 282 unlabeled videos (6 million frames). For more details, we refer the reader to the dataset reference (Sun *et al.*, 2021). The coordinates were recorded at a rate of 30 frames per second.

The CRIM13 dataset (Burgos-Artizzu *et al.*, 2012) consists of 88 hours of annotated videos. As the original dataset only provides tracking information for the body center of the animals, videos were re-tracked using DeepLabCut (Mathis *et al.*, 2018), following the same seven body parts configuration in the CalMS21 dataset (Sun *et al.*, 2021). Furthermore, videos were re-organized based on the metadata of the experiments (i.e. intruder sex, mouse line and experimental conditions) to allow group comparisons in this study. Videos were curated to select only conditions without specific treatment as experimental conditions (e.g. anesthetized or ovariectomized were excluded), with BALB/c line used as intruder and segments of recording successfully tracked containing two animals for at least 5 minutes. Where available, human annotations were verified and synchronized with the tracking data.

The VTA dataset was acquired in house using recorded videos of free social interactions and previously published recordings (Solie, Girard *et al.,* 2024; https://zenodo.org/record/5564893). Briefly, wild-type C57BL/6J mice and transgenic DAT-Cre mice were employed. VTA DA neuron recordings were made using Neuralynx and a custom multi-unit recording microdrive with octrodes coupled with optic fiber. Surgical procedures involved viral injection of rAAV5-Ef1α-DIO-hChR2(H134R)-eYFP into the VTA, followed by microdrive implantation at following stereotactic coordinates: ML ± 0.5 mm, AP −3.2 mm, DV −4.20 ± 0.05 mm from bregma. Optogenetic photolabeling was used to validate the dopaminergic nature of recorded neurons. The mice performed a free social interaction task, interacting freely with an unfamiliar conspecific. In vivo recording data was analyzed using custom MATLAB code for spike sorting, feature extraction and classification of VTA DA neurons. Body-pose estimation was extracted using DeepLabCut (Mathis *et al.*, 2018; Lauer *et al.*, 2022), following the same seven body parts configuration in the CalMS21 dataset (Sun *et al.*, 2021).

## Transformer network architecture

The model backbone is adapted from the ViT architecture (Dosovitskiy *et al.*, 2020), akin to the factorized encoder proposed in ViViT (Arnab *et al.*, 2021). The frame encoder is a multi-layer perceptron (MLP) with GELU (Gaussian Error Linear Unit) activations. Positional encodings were learned during the training process and added to the frame embeddings.

Unlike ViT/ViViT, no special tokens were used to represent the class label or to separate different portions of the input in the transformer encoder. These tokens were not necessary as the number of body coordinates and window length are fixed. The function of the class token was implemented by adding a Max Pooling layer before the classification heads if required.

Classification heads were also implemented as MLPs. No dropout layers or other loss regularization techniques were used. All MLPs in the model have the same activation functions and hyperparameters. Unless otherwise stated, the window size was 200 frames.

## Model fitting

The model was trained using four self-supervised learning tasks: Swap Mouse Prediction (SMP), Next-Window Prediction (NWP), Video Speed Prediction (VSP), and Delay Mouse Prediction (DMP). These tasks were defined as binary classification problems (i.e., original sequence vs altered sequence). At each training step, one example for each task was presented to the model backbone to compute the corresponding LISBET embedding and classified using a task-specific head. The backbone weights were shared across tasks. Model performance was calculated as binary accuracy. Label smoothing was used to improve generalization. The number of training epochs was determined during hyperparameter tuning.

Models for frame classification were either fine-tuned from a pre-trained LISBET embedding model (self-supervised) or trained from scratch as control cases. Frame classification was performed using a linear decoder and evaluated in terms of F1 score. For the CalMS21 dataset, the "other" class was excluded from the score calculations, as suggested by its authors (Sun *et al.*, 2021). Model evaluation was always performed on a held-out test set never used during training or hyperparameter tuning.

It should be noted that, at each epoch, the training and validation set are randomly generated from the source training and validation data. This implies that every epoch is unique compared to the actual input data, although the source sequences (i.e., body pose estimation data) used for each set are frozen and no data spillover is allowed.

## Hyper parameters tuning

The main hyperparameters of the transformer model were chosen via a custom grid search:
- Embedding dimension in [16, 32, 64, 128],
- Number of layers and encoder heads in [2, 4, 8, 16],
- MLP hidden layer dimension in [512, 1024, 2048, 4096].

To reduce the computational cost of the search, 12 model configurations were chosen from the grid with a progressively increasing number of parameters. Each candidate was evaluated on 4 cross-validation splits of the training set (repeated random sub-sampling validation, 90/10 ratio of training over validation data).The model was trained for 100 epochs, as described in section "Model Fitting", using the CalMS21 unlabeled dataset (Sun *et al.*, 2021). The performance of each configuration was assessed as the mean accuracy over the last 10 training epochs of the corresponding models across training tasks and cross-validation splits. The model with the highest score was chosen as the best candidate and re-trained on the whole training set. Mixed-precision (FP16) to train models exceeding GPU memory constraints (i.e., MLP hidden layer dim > 2048).

## Motifs segmentation

Motifs were segmented using a Gaussian Hidden Markov Model (Gaussian-HMM), fitted using the expectation-maximization (EM) algorithm. Model fitting was halted after convergence (delta log-likelihood < 0.01) or 500 EM steps.

## Prototype selection

A range of Hidden Markov Models (HMMs) from 2 to 32 states was applied to generate Motifs. All Motifs were compared using the Jaccard index and hierarchically clustered. Silhouette scores determined the optimal number of clusters. Within each macro-category, the motif with the highest silhouette score was identified as the prototype.

## Behavioral analysis

For the VTA neuronal recording dataset, behavioral features were computed from video tracking data to characterize the prototypes obtained from LISBET. Proximity was obtained from Euclidean distance between the experimental animal's body center and the stimulus animal's snout. Orientation was obtained from angle between vectors from body center to snout for both animals in degrees. Velocity was obtained from instantaneous velocity of the both animal's body center (cm/sec). Peri-Event Time Histograms were obtained by aligning these features to the start of each prototype event (±5 sec). Illustrative videos were created by concatenating segments (3 sec before to 3 sec after prototype start) for randomly selected events of each prototype presented. Events shorter than 200ms or longer than 2sec were discarded. For the CalMS21 dataset, illustrative videos were created by concatenating segments (1 sec before prototype start to 1 sec after prototype end) for randomly selected events of each prototype presented.

## Statistics

Statistical analysis was conducted using scipy. The number of experiments (n) is indicated in the figure legends. Statistical analyses for male versus female in Fig. 5 were performed with an unpaired t-test. Unless otherwise stated, summary data are presented in figures as boxplot.

## Data availability

The CalMS21 and CRIM13 datasets are available online on the website of their developers. The VTA dataset is available upon request to the authors. Analysis scripts and results will be publicly available on Zenodo in the upcoming weeks.

## Code availability

LISBET source code, documentation, examples and weights of the best-performing models are available in closed beta testing upon request to the authors, and will be made publicly available in the upcoming weeks at https://github.com/BelloneLab/lisbet.

# Figures

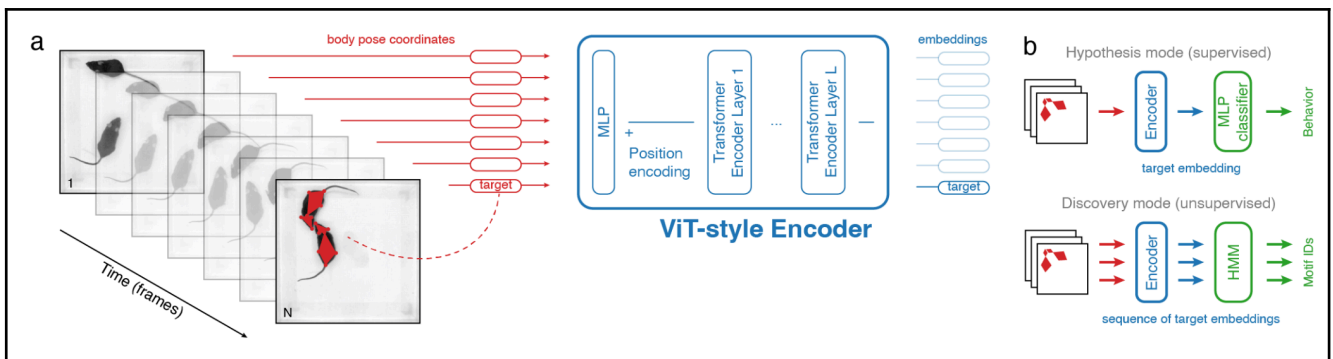

**Fig. 1: Schematic of LISBET architecture and operation modes. a**, Body pose coordinates from behavioral videos are analyzed using a sliding window of user-prescribed size (N frames). For each window, one frame is chosen as target, while all the others are provided as context. Windows are processed through a ViT-style encoder to produce a set of learned behavioral features for each frame (embeddings). **b**, Embeddings can then be used for behavior classification (supervised, hypothesis mode) or motif identification (unsupervised, discovery mode). In hypothesis mode (top), the embedding corresponding to the target frame is assigned to a behavior using a Multi-Layer Perceptron (MLP) classifier, trained on user-provided annotations. In discovery mode (bottom), the embeddings corresponding to the target frames are accumulated for the whole video sequence and segmented into behavioral motifs using a Hidden Markov Model (HMM).

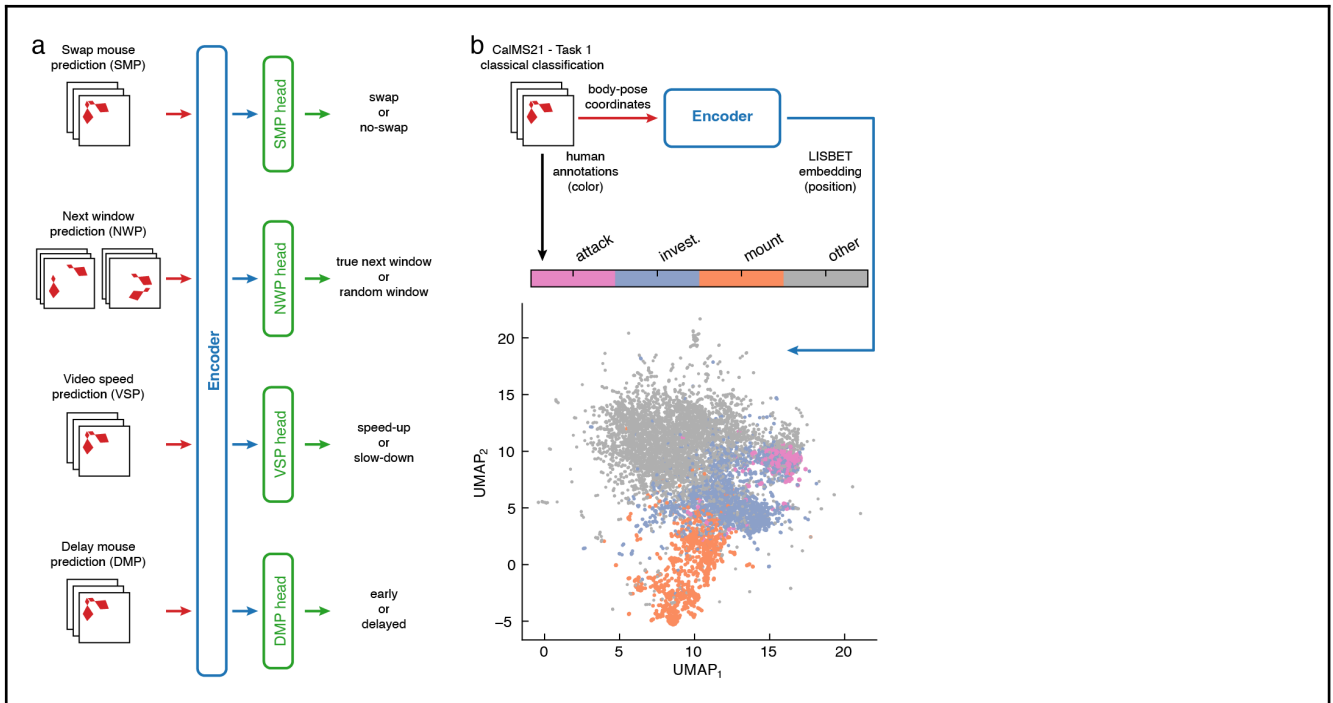

**Fig. 2: Self-supervised training and comparison with human annotations. a**, Schematic of the self-supervised training strategy. Four training tasks are concurrently solved using a shared LISBET backbone and four, task-specific, classification heads. Each head is composed of a Max Pooling Layer followed by a MLP Layer. After training, the classification heads are discarded and only the LISBET embedding model is kept for subsequent analysis. **b**, Schematic of analysis pipeline and visualization of the LISBET embedding in reduced dimension using UMAP after the self-supervised training (frozen weights). The data represented is a random sample (n = 10000) of the test set in the CalMS21 - Task 1 dataset (Sun *et al.*, 2021). The position of dots corresponds to LISBET embedding obtained from the time windows analyzed and color overlay corresponds to the independent human annotations.

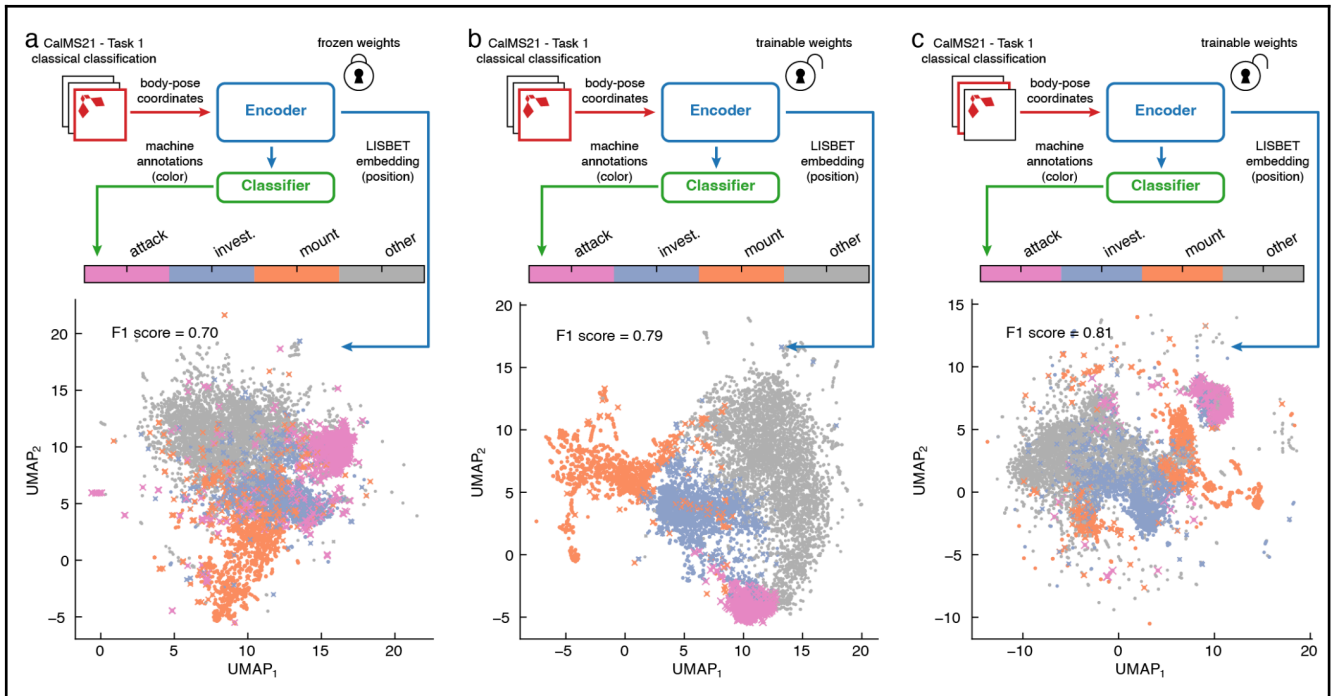

**Fig. 3: Supervised fine-tuning to reproduce human annotations. a**, Same as in Fig. 2b, but for the hypothesis mode pipeline. An MLP classifier is fitted on the frozen LISBET embeddings to reproduce the human annotations in the training set in the CalMS21 - Task 1 dataset (Sun *et al.*, 2021). **b**, Same as in **a**, but the LISBET embeddings are fine-tuned together with the MLP classifier. **c**, Same as in **b**, but using the central frame of the window as target. Embeddings and the machine annotations shown here are from the test set. Dots represent correctly classified examples, while crosses are the misclassified ones. Random sample size n = 10000 in **a-c**.

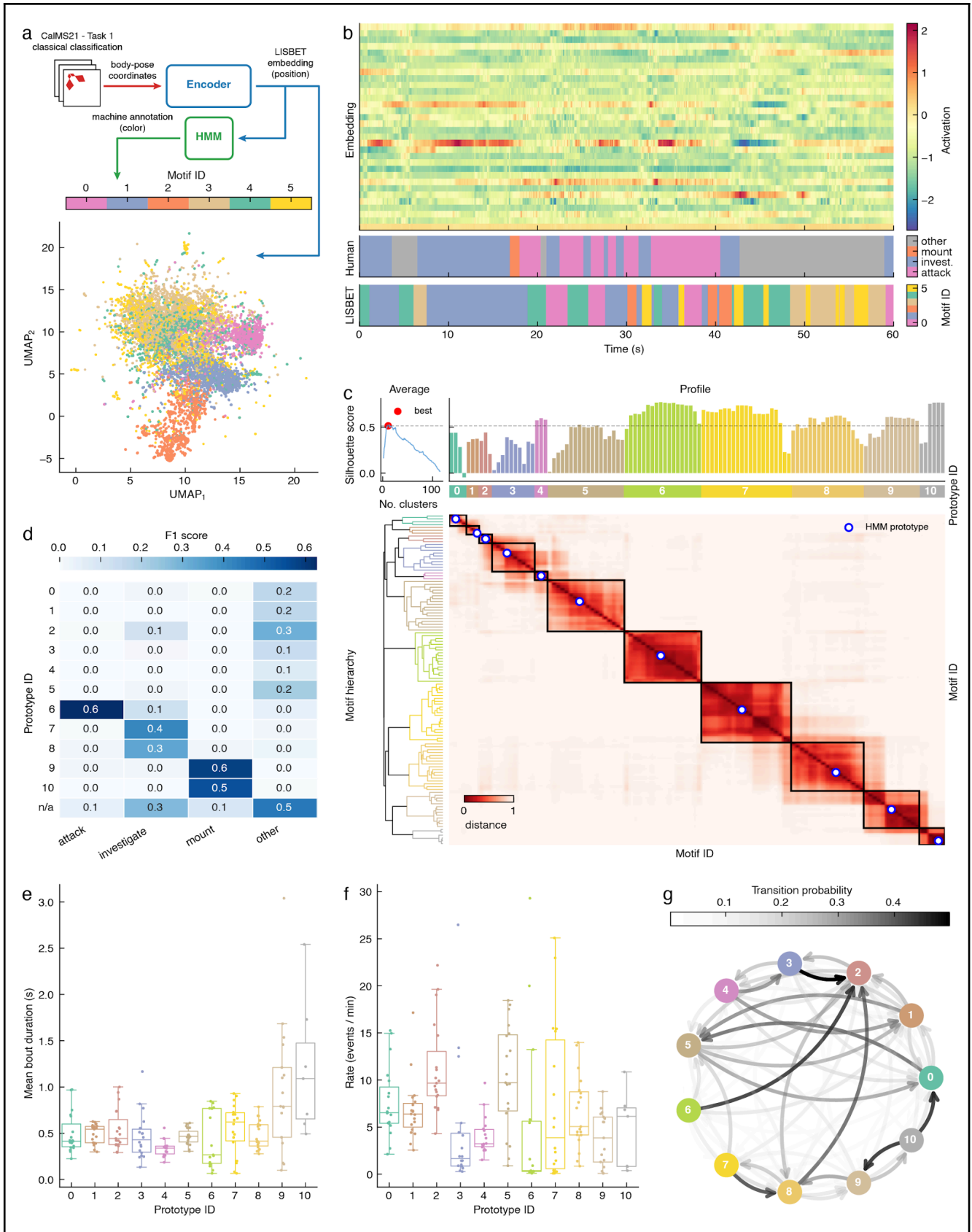

**Fig. 4: Unsupervised embedding segmentation into behavioral motifs. a**, Schematic of analysis pipeline in discovery mode via Hidden Markov Models (HMMs, top) and visualization of the LISBET embeddings in reduced dimension using UMAP (bottom). The data represented is a random sample (n = 10000) of the test set in the CalMS21 - Task 1 dataset (Sun *et al.*, 2021). The position of dots corresponds to LISBET embedding obtained from the time windows analyzed (as Fig. 1b) and color overlay corresponds to social motifs obtained from automatic segmentation of LISBET embedding using HMM with 7 hidden states. **b**, Example of a video segment of 60 seconds

showing temporal alignment of corresponding heatmap of LISBET embeddings (activation value of last layer of LISBET backbone, top), with corresponding human annotations (middle) and social motifs obtained from segmentation of LISBET embedding with the HMM (bottom). LISBET annotations in **a** and **b** were post processed using a causal median filter (filter size = 1 s). **c**, Automatic motif selection. Top left: silhouette scores used to determine optimal clusters of motifs macro-categories. Bottom left: hierarchical motif clustering. Top right: Selection of prototype motifs within each macro-category. Bottom right: motifs distance matrix (Jaccard distance) with the identified macro-categories delineated by black edge square with corresponding prototype motif (blue circles). **d**, Coverage of human-annotated behaviors by LISBET prototypes using the F1 score. **e**, Distribution of prototype mean duration (seconds). Each dot represents the mean duration of the corresponding prototype in one complete sequence from the test set (n = 19). **f**, Distribution of prototypes rate (events per minute). Each dot represents the mean rate of the corresponding prototype in one sequence from the test set (n = 19). **g**, Transition probability between prototypes. Sample size **e-f** n = 19 for p0, p1, p2, p5, p8; n = 18 for p3, p4, p7; n = 17 for p9; n = 15 for p6; n = 7 for p10.

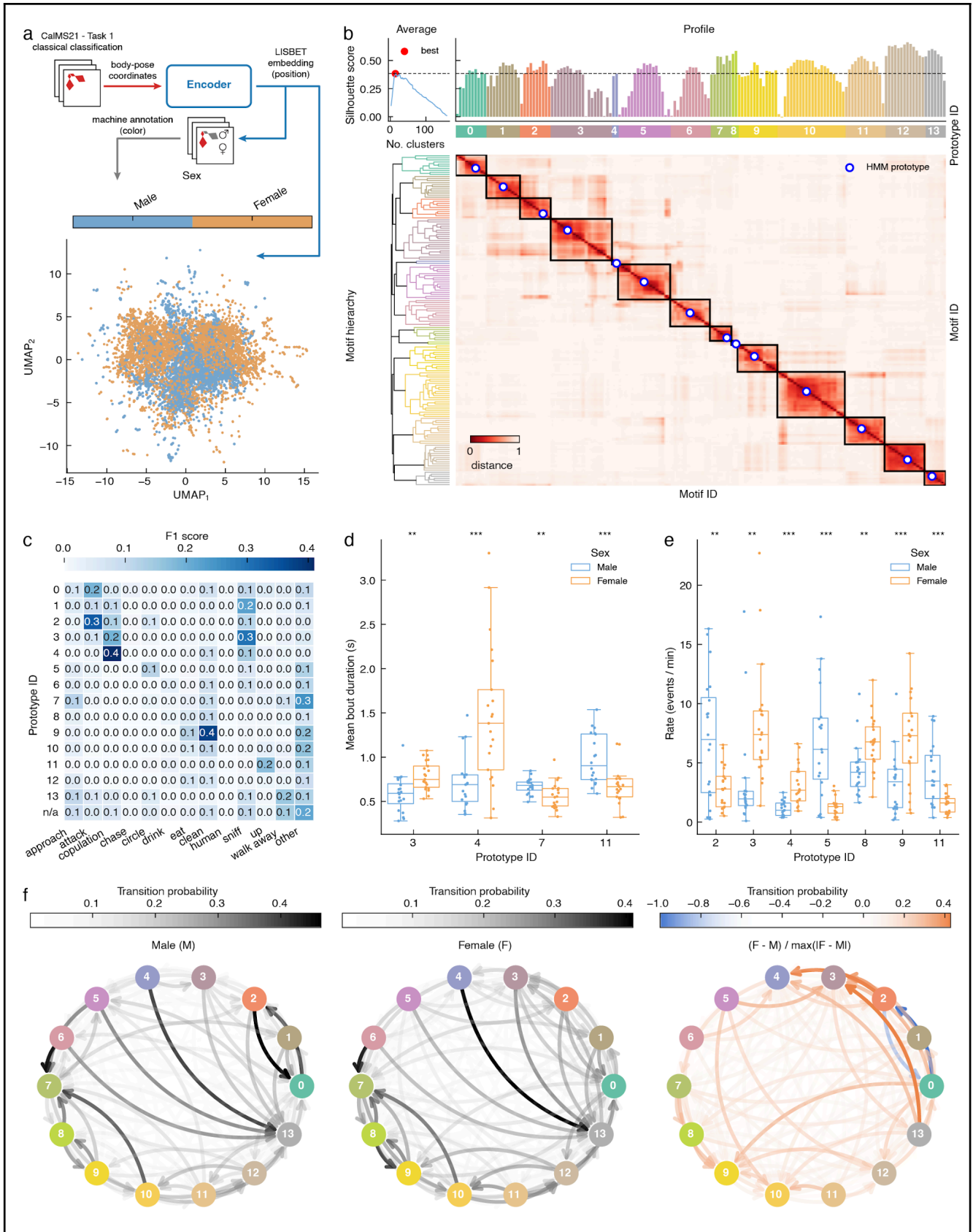

**Fig. 5: Unsupervised behavioral phenotyping. a**, Schematic of analysis pipeline in discovery mode via Hidden Markov Models for group comparisons (HMMs, top) and visualization of the LISBET embedding in reduced dimension using UMAP (bottom). The data represented is a random sample (n = 10000) of the CRIM13 dataset (Burgos-Artizzu *et al.*, 2012). The position of dots corresponds to LISBET embedding obtained from the time windows analyzed (as Fig. 2b) and color overlay corresponds to the sex of the intruder mouse in the experiments. **b**, Automatic motif selection, as in Fig. 3c. **c**, Coverage of human-annotated behaviors by LISBET prototypes using

the F1 score, as in Fig. 3d. **d**, Distribution of significantly different (p < 0.01) prototype mean durations (seconds) between male (blue) and female (orange) intruder mice. Each dot represents the mean duration of the corresponding prototype in one complete sequence from the data set. **e**, Distribution of significantly different (p < 0.01) prototypes rate (events per minute) between male (blue) and female (orange) intruder mice. Each dot represents the mean rate of the corresponding prototype in one sequence from the test set. **f**, Transition probability between social motifs in the male vs male case (left), male vs female (middle), and relative difference between them (right). Sample size for **d-e** n = (20 male, 21 female) for all prototypes, except n = (19 male, 21 female) for p4 and n = (19 male, 18 female) for p5.

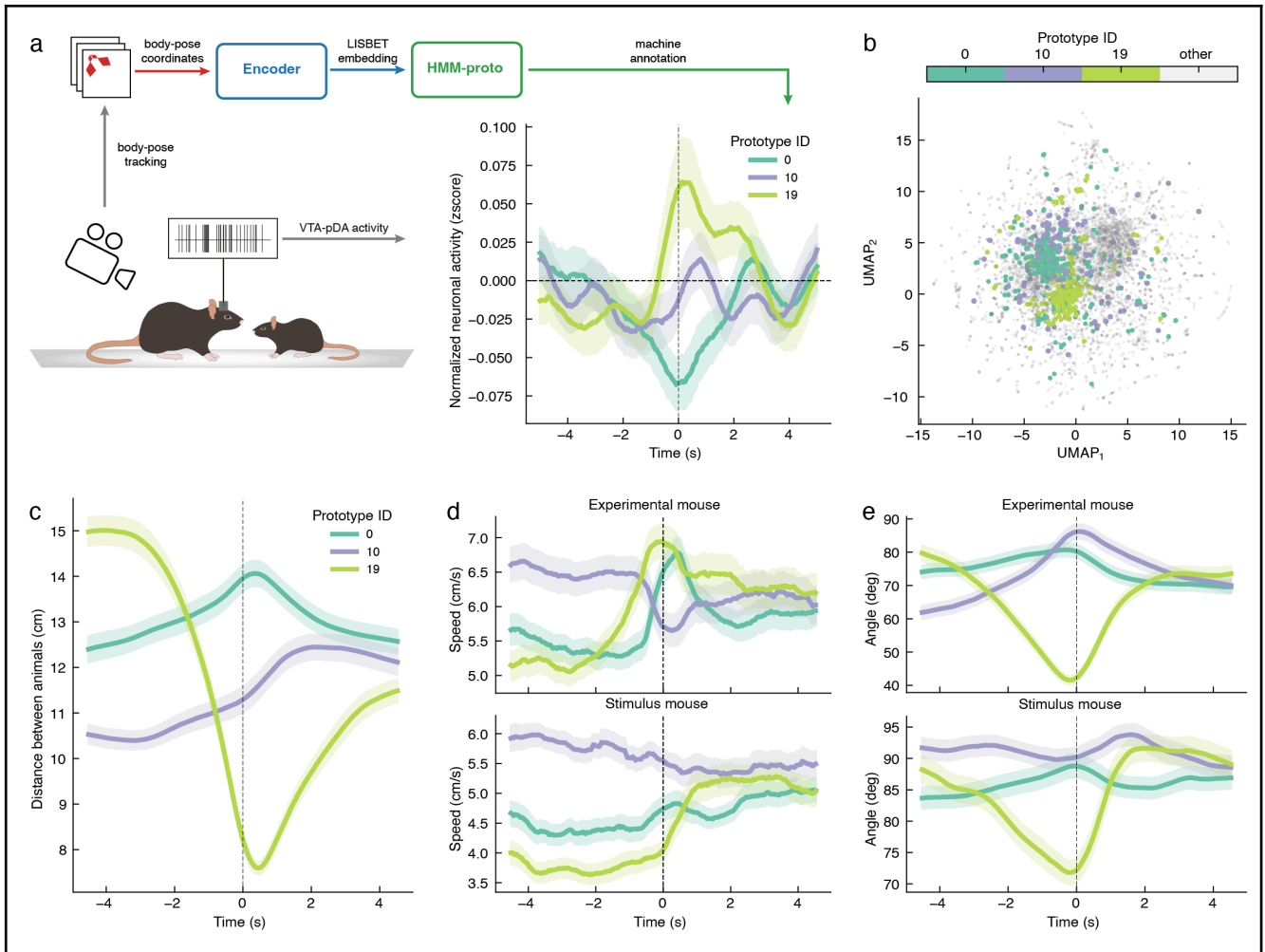

**Fig. 6: Unsupervised identification of neural correlates of social behavior in the VTA. a**, Schematic of analysis pipeline (top and left) and Perievent Time Histogram (PETH) of the normalized firing rate of VTA-pDA neurons for three exemplar social prototypes (right). Sample size n = 134 for every prototype; Gaussian smoothing (window size = 1 s). **b**, Visualization of the LISBET embeddings in reduced dimension using UMAP for the three exemplar prototypes shown in **a** (n = 10000, random sample). **c**, Distance between animals during the exemplar prototypes. **d**, Speed of animals (experimental mouse, top; stimulus mouse, bottom) during the exemplar prototypes. **e**, Angle between head of animals (experimental mouse, top; stimulus mouse, bottom) during the exemplar prototypes. In **a** and **c-e**, solid lines represent the mean, while shaded areas the SEM. Sample size for **c-e** n = 100 (p0 and p10), 99 (p19). Raw traces used for computing mean and SEM in **c-e** were filtered using a moving average window (window size = 1 s).

# Extended Data Figures

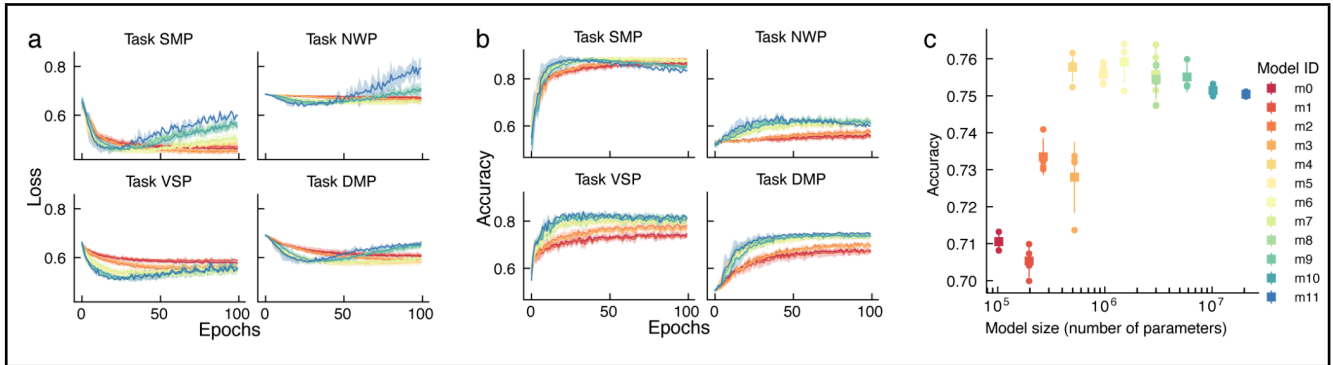

**Extended Data Fig. 1: LISBET tuning results. a,** Evolution of dev set losses during training for each self-supervised task. Solid lines represent the mean loss across cross-validation folds while the shaded areas represent the corresponding standard deviation. Model ID color codes as in **c. b,** Evolution of dev set binary accuracy during training for each self-supervised task. Solid lines and shaded areas have the same meaning as in **b,** Model ID color codes as in **c. c,** Binary accuracy summary. Circles represent the mean of the last 5 epochs in each cross-validation fold. Squares and vertical lines represent the mean and standard deviation across cross-validation folds respectively. Sample size n = 4 for every model, except model m9 with n = 3 and model m11 with n = 2.

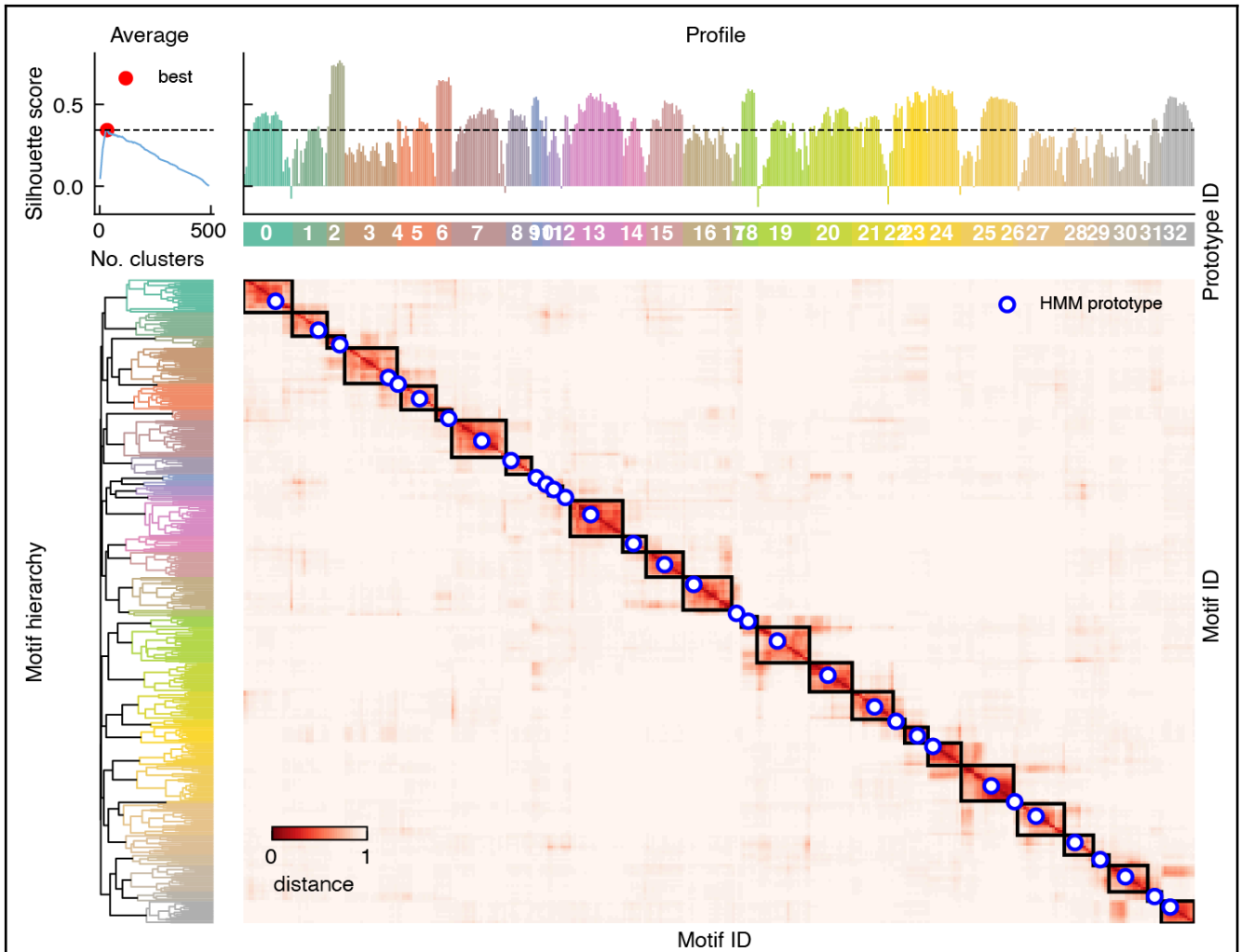

**Extended Data Fig. 2: Automatic motif selection for the VTA-pDA dataset.** Top left: silhouette scores used to determine optimal clusters of motifs macro-categories. Bottom left: hierarchical motif clustering. Top right: Selection of prototype motifs within each macro-category. Bottom right: motifs distance matrix (Jaccard distance) with the identified macro-categories delineated by black edge square with corresponding prototype motif (blue circles) and human annotations.

# Tables

| Encoder pre-training | Encoder fine-tuning | Window offset | F1 score | NMI |
|---|---|---|---|---|
| True | True | 0 | 0.794852 | 0.636776 |
| False | True | 0 | 0.733333 | 0.567618 |
| True | False | 0 | 0.699080 | 0.504681 |
| | True | 49 | 0.804781 | 0.637817 |
| | | 99 | 0.812269 | 0.646746 |
| | | 149 | 0.802673 | 0.637473 |
| | | 199 | 0.785222 | 0.623323 |

**Table 1: Calm21 Task 1 results.**

# Acknowledgements

The authors would like to acknowledge Alexandre Pouget and Charles Findling for helping conceptualize the Next Window Prediction (NWP) task and for discussions on the unsupervised modeling process. Bastien Redon, Ivan Rodriguez and Leon Fodoulian for suggestions on the manuscript. Mackenzie Mathis for discussions on the project. Andrea Della Valle, Thomas Maillard, Thibaut Chataing, Laura Schwarz and Nikoloz Sirmpilatze for feedback and suggestions on the LISBET implementation.

The computations were performed at University of Geneva using Baobab HPC service.

During the preparation of this work the authors used large language models (OpenAI ChatGPT and Anthropic Claude) for language enhancement. These tools and services did not contribute to the scientific content and the authors retain sole responsibility for the content, interpretations, and conclusions.

# Contributions

G.C., B.G., and C.B. conceived the study. G.C. and B.G. conceptualized the machine learning model. G.C. implemented and optimized the ML model. G.C. performed the computational experiments. B.G. performed the in vivo experiments and extracted body pose coordinates from the videos. G.C. and B.G. conceptualized and implemented the analysis pipelines. G.C., B.G., and C.B. wrote the manuscript.